\documentclass{llncs}
\usepackage{float}
\usepackage{graphicx} % Bilder
\usepackage{color} % Farben
\definecolor{brickorange}{RGB}{193,74,9}
\definecolor{blue2}{RGB}{162,191,254}
\definecolor{red}{RGB}{255,0,0}
\definecolor{green}{RGB}{0,255,0}
\usepackage{caption} % Verbesserte Untertitel
\usepackage{subfigure} % mehrere Abbildungen nebeneinander/übereinander
\usepackage{wrapfig} % wrap figures in text
\usepackage{amsmath,amssymb,marvosym} % Math
\usepackage[table]{xcolor} % add colors to tables

\usepackage{booktabs} %change the tickness of table
\usepackage[colorinlistoftodos]{todonotes} %% add comments
\usepackage{algorithm} % describe algorithms
\usepackage[noend]{algpseudocode}

\usepackage{pgfplots}
\usepackage{environ}
\makeatletter
\newsavebox{\measure@tikzpicture}
\NewEnviron{scaletikzpicturetowidth}[1]{%
  \def\tikz@width{#1}%
  \begin{lrbox}{\measure@tikzpicture}%
  \BODY
  \end{lrbox}%
  \pgfmathparse{#1/\wd\measure@tikzpicture}%
  \BODY
}
\makeatother
\usepackage{enumitem}
\usepackage{makecell}
\usepackage{url}
\usepackage{wasysym}
\usepackage{multirow}

\usepackage{tikz}
 
\newcommand*{\tikzbullet}[2]{%
   \setbox0=\hbox{\strut}%
   \begin{tikzpicture}
     \useasboundingbox (-.25em,0) rectangle (.25em,\ht0);
     \filldraw[draw=#1,fill=#2] (0,0.5\ht0) circle[radius=.25em];
   \end{tikzpicture}%
}

\begin{document}

\title{Webly Supervised Learning for Skin Lesion Classification}
\titlerunning{Fernando Navarro et al.}
% \titlerunning{Anonymous}  % abbreviated title (for running head)
%                                     also used for the TOC unless
%                                     \toctitle is used
%
\author{Fernando Navarro\inst{1} \and Sailesh Conjeti\inst{1,2} \and
Federico Tombari\inst{1} \and Nassir Navab\inst{1,3}}
% \author{Paper \# 224}
%

%
%%%% list of authors for the TOC (use if author list has to be modified)
% \tocauthor{Anonymous}
%
% \institute{Under Review for MICCAI 2018}
\institute{Computer Aided Medical Procedures, Technische Universit\"at M\"unchen, Germany\\
%\email{fernando.navarro@tum.de}
\and
German Center for Neurodegenrative Diseases (DZNE), Bonn, Germany \\
\and
Computer Aided Medical Procedures, Johns Hopkins University, Baltimore,USA}

\maketitle              % typeset the title of the contribution

\begin{abstract}

Within medical imaging, manual curation of sufficient well-labeled samples is cost, time and scale-prohibitive. To improve the representativeness of the training dataset, for the first time, we present an approach to utilize large amounts of freely available web data through web-crawling. To handle noise and weak nature of web annotations, we propose a two-step transfer learning based training process with a robust loss function, termed as Webly Supervised Learning (WSL) to train deep models for the task. We also leverage \textit{search by image} to improve the search specificity of our web-crawling and reduce cross-domain noise. Within WSL, we explicitly model the noise structure between classes and incorporate it to selectively distill knowledge from the web data during model training. To demonstrate improved performance due to WSL, we benchmarked on a publicly available 10-class fine-grained skin lesion classification dataset and report a significant improvement of top-1 classification accuracy from 71.25 \% to 80.53 \% due to the incorporation of web-supervision.

\end{abstract}
\section{Introduction}

The success of deep learning in computer vision tasks such as image classification, object detection, segmentation \textit{etc.} is owed to the availability of a large corpus of annotated training data~\cite{resnet,inceptionv3,inceptionv4}. However, translating these developments to medical imaging applications is often challenging as curating a representative dataset is cost-, time- and scale-prohibitive. On the other hand, excessive reliance on a small-sized, well-curated dataset offers limited guarantees on the generalizability to unseen scenarios and could lead to potential overfit on the training data due to excessive over-parameterization of deep networks. In this paper, we propose to leverage freely available data crawled from the web to offset the need for a large dataset and introduce the concept of \textit{Webly Supervised Learning} (WSL) as a potential approach for training neural networks for medical imaging applications. We present a proof of concept for the task of fine-grained classification of skin lesions in dermatological images. 

\begin{figure}
\centering
\begin{minipage}{0.5\textwidth}
\begin{center}
\includegraphics[width=\textwidth]{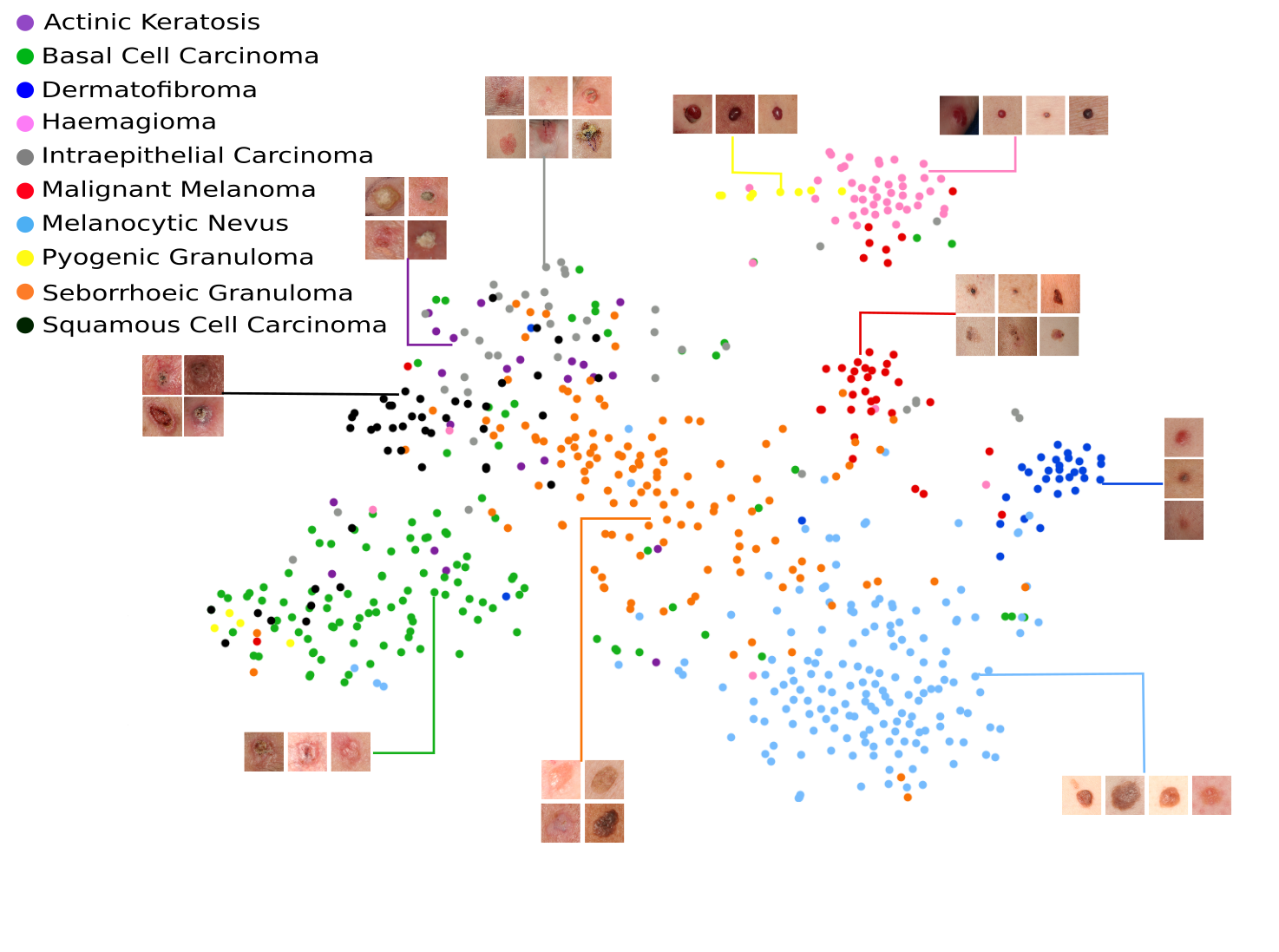}
\end{center}
% \caption*{\textbf{CleanNet}}
\end{minipage}\hfill
\begin{minipage}{0.5\textwidth}
\begin{center}
\includegraphics[width=\textwidth]{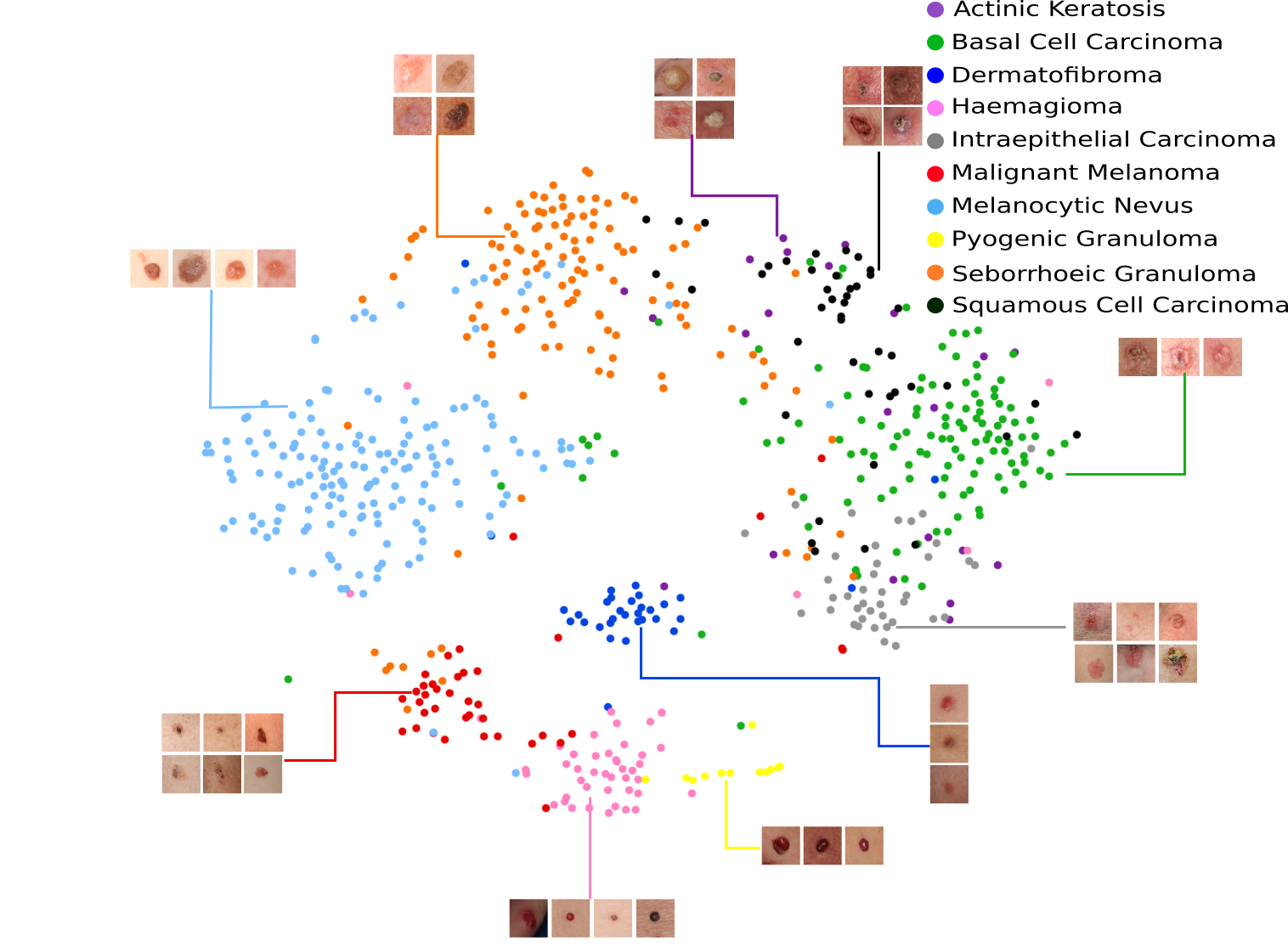}
\end{center}
% \caption*{\textbf{LocalWebNet}}
\end{minipage}
\vspace{-5pt}
\caption{\small{Comparison of t-SNE embedding space generated from networks trained on limited clean data (Left) against network trained with Webly Supervised Learning (Right) generating compact class clusters with improved separability especially for under-represented classes.}}
\label{fig:embedding}
\vspace{-15pt}
\end{figure}
% \vspace{-10pt}
The task of skin lesion classification is a representative example of a medical imaging application in which, annotated training data is limited in availability. However, there is an abundance of freely-available web data. We source our images from multiple publicly-accessible sites such as~\cite{webdata2}, where pictures of skin lesions are uploaded with the goal of getting feedback on the type of lesion with respect to visual features. Prior work on deep learning for skin lesion classification includes training networks that perform either a two or three-class classification (melanoma, non-cancerous and seborrheic keratosis)~\cite{tmi1,isicpaper1,isbi1}. Authors in~\cite{naturepaper} propose a deeply learned network for nine-class categorization using a large dataset of 130000 images extensively curated from hospital archives and from dermatological websites. The data used in this work underwent extensive manual quality control with 23 human experts and filtering prior to fine-tuning InceptionV3~\cite{inceptionv3}. In contrast to~\cite{naturepaper}, within this paper, we adopt a more unconstrained learning paradigm by focusing on learning in presence of extreme label noise by developing a dedicated robust loss function and employing transfer learning strategies to seamlessly leverage webly sourced data into training without employing any additional heuristics or expert knowledge.
\begin{wrapfigure}[11]{l}{0.5\textwidth}
\vspace{-25pt}
\centering
\includegraphics[width=0.49\textwidth]{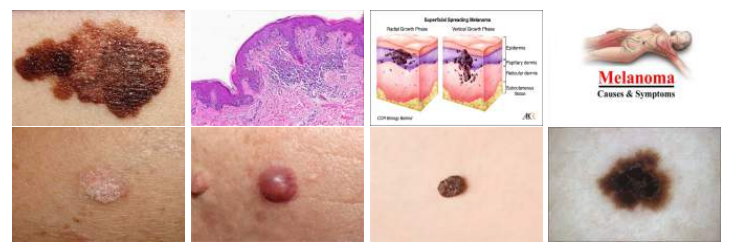}
\caption{\footnotesize{Type of noise in WSL for Melanoma class as keyword. The images in the first row represent examples of cross-domain noise. The second row represents the cross-category noise.}}
\label{fig:noisetype}
\end{wrapfigure}Harvesting images from the web presents opportunities for abundant availability and the ability to encompass sufficient heterogeneity during training. However, learning from them is challenging due to the presence of different types of noise. These include \textit{cross-domain noise}: retrieved web images may include non-dermatoscopic images such as histology, artistic illustrations, icons \textit{etc.} and \textit{cross-category noise}: images that are visually similar to the query yet belong to a different class. Cross-domain noise is introduced by bias due to a specific search engine and associated search criterion (such as user tags).
\begin{figure}[ht]
\begin{center}
\includegraphics[width=\textwidth]{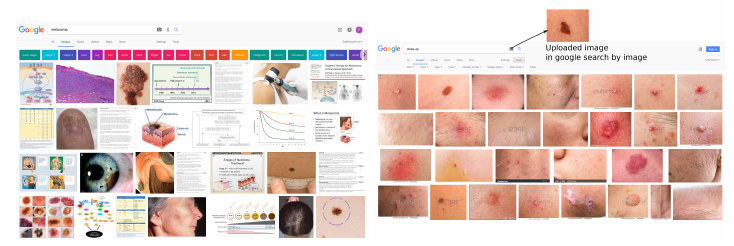}
\end{center}
\vspace{-10pt}
\caption{\small{Comparison of crawled results. The left image shows an example of a search by keyword ``melanoma": the resulting images contain high cross-domain noise. The right image shows the results of a search by image, where the cross-domain noise is significantly reduced sharing strong visual similarity to the query image.}}
\label{fig:searchkeyword}
\vspace{-10pt}
\end{figure}
% \begin{wrapfigure}{r}{0.5\textwidth}
% \centering
% \includegraphics[width=0.5\textwidth]{images/search_comparison2}
% \caption{The left image shows an example of a search by keyword "melanoma": the resulting images contain high cross-domain noise. The right image shows the results of a search by image}
% \label{fig:searchkeyword}
% \end{wrapfigure}
Additionally, image-search engines are biased as they often operate in high-precision low-recall regimes and preferentially present objects centered with clean background and a canonical view-point. Fig.~\ref{fig:noisetype} illustrates different types of noise present in retrieved images upon web-crawling with "melanoma" as the search tag. Learning from web data is one approach for learning under extreme label noise. Methods within the computer vision community that leverage web supervision for training can be broadly categorized as: (1) \textit{Filtering}: approaches that aim to clean or filter the collected web images prior to training~\cite{clean1,clean2}; (2) \textit{Modeling Relationships}: approaches that model the relationship between web images and noisy labels with a small subset of clean images and utilize the discovered relationships to improve training~\cite{graphical1,graphical2} and (3) \textit{Robust Loss Functions}: approaches that learn in the presence of label noise by introducing robustness within their loss-function design \cite{robust1,robust2}. Our proposed approach encompasses the best of the aforementioned approaches with the following contributions: 
\begin{enumerate}[topsep=0pt,itemsep=-1ex,partopsep=1ex,parsep=1ex]
\item \textbf{Reduction in Cross-domain Noise}: This is the first work to leverage \textit{search by image} to improve search specificity and reduce cross-domain noise by fetching images that share close visual features to the query image.
\item \textbf{Noise Modeling}: We model the noise as a class-transition matrix which is estimated from the bag of retrieved images. This noise modeling approach allows for distillation of knowledge from noisy web-images to train very-deep networks. 
\item  To the best of our knowledge, this is the first work within the medical image computing that leverages web-supervision to train deep neural networks and specifically targeted at fine-grained ten-class categorization of skin lesions.
\end{enumerate}

\section{Methodology}
% \begin{figure}[t]
% \begin{center}
% \includegraphics[width=0.49\textwidth]{images/search3}
% \includegraphics[width=0.49\textwidth]{images/searchimage}
% \end{center}
% \caption{The left image shows an example of a search by keyword "melanoma": the resulting images contain high cross-domain noise. The right image shows the results of a search by image}
% \label{fig:searchkeyword}
% \end{figure}
\begin{figure}[ht]
\begin{center}
\includegraphics[width=0.8\textwidth]{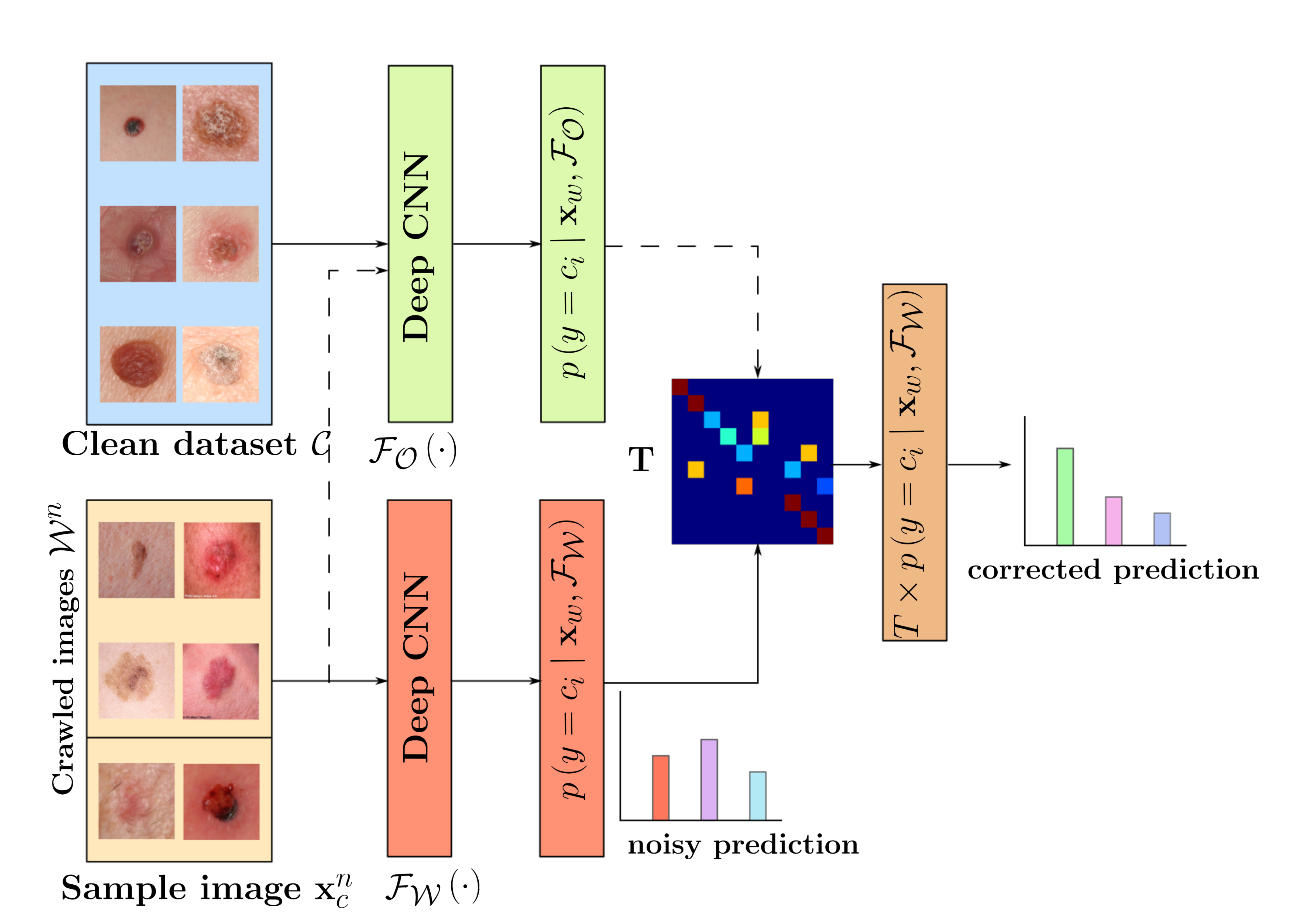}
\end{center}
\vspace{-10pt}
\caption{\small{Overview of the proposed WSL approach consisting of a two-step  training. First, training a network on web data, follow by fine-tuning a second network utilizing the latter as strong prior initialization. Noise correction is performed when training on web data.}}
\label{fig:overview}
\vspace{-0.68cm}
\end{figure}

% \subsection{Web-Crawling}
% \begin{wrapfigure}[16]{r}{0.5\textwidth}
% \centering
% \includegraphics[width=0.49\textwidth]{images/overview_latex_final}
% \caption{Overview of the proposed WSL approach.}
% \label{fig:overview}
% \end{wrapfigure}
\vspace{-10pt}
Given a small representative training dataset $\mathcal{C} = \left \{ \left ( \mathbf{x}_{c}^{n}, \mathbf{y}_{c}^{n} \right ) \right \}_{n=1}^{N}$ of dermatological images with expert annotations, we source web-images for WSL by utilizing the \textit{Search by Image} option within standard search engines (here, \url{https://images.google.com/}) by submitting each of the \textit{clean} images independently and crawling the top retrieved results. Let the bag of images (of size $M$) crawled from the web for a query image of $\mathbf{x}_{c}^{n}$ be represented as $\mathcal{W}^{n} = \left \{  \mathbf{x}_{w}^{m} \mid \text{Query: }\mathbf{x}^{n}_{c} \right \}_{m=1}^{M}$. The semantic label associated with the query clean image $\mathbf{y}_{c}^{n}$ is given to the corresponding web-crawled bag $\mathcal{W}_{n}$. Let the complete web-crawled images with their corresponding annotations be denoted as $\mathcal{W} = \left \{ \left ( \mathcal{W}^{n}, \mathbf{y}_{c}^{n} \right ) \right \}_{n=1}^{N}$. Contrasting with prior WSL approaches~\cite{webly1,webly2} that use search by keyword (such as \textit{melanoma}, \textit{keratosis} \textit{etc.}), we observe that our approach significantly reduces the cross-domain noise by fetching only images that share strong visual features with the query. Fig.~\ref{fig:searchkeyword} contrasts the proposed search by image approach against the search by keyword methodology for the construction of a web-dataset for WSL. It must be noted that the labels transferred from $\mathcal{C}$ to $\mathcal{W}$ are extremely noisy as the web-search relies on non-task specific solely visual features for ranking and carries no guarantees on the fetched results sharing the same semantic class as the query. Additionally, in such an uncontrolled setting, multiple queries could fetch the same images thus the web-images could carry potential cross-category noise. With per-image web-crawling, our training dataset is significantly augmented (with at most $M \times N$ unique images) and the resultant dataset is rich in representativeness and heterogeneity but fraught with extreme label noise which needs to be factored out in the subsequent training steps.  

\subsection{Model Learning}

\noindent
\textbf{Noise Correction}: Assuming that we have access to a perfect oracle network $\mathcal{F}_{\mathcal{O}}\left ( \cdot \right )$, we model the noise within the web images as a class-transition matrix $T$ that can be used to diffuse the predictions on web data across confusing classes. In na\"{i}ve terms, $T$ models the probability of each class being confused into one another. Considering three classes ($c_{1},c_{2}$ and $c_{3}$), if $c_{1}$ and $c_{2}$ are visually more similar than $c_{3}$, there is a higher probability for cross-category noise across $c_{1}$ and $c_{2}$ in comparison to $c_{1}$ and $c_{3}$ (or $c_{2}$ and $c_{3}$) and this reflects back in the estimated class-transition matrix as $T\left (c_{1},c_{2}  \right ) > T\left (c_{1},c_{3}  \right )$. From within the web crawled images $\mathcal{W}$, we use the predictions of oracle network  $\mathcal{F}_{\mathcal{O}}$ to mine the most representative sample $\hat{\mathbf{x}}_{w,c_{i}}$ of class $c_{i}$ from within $\mathcal{W}$ as: 
\begin{equation}
\hat{\mathbf{x}}_{w,c_{i}} = \text{argmax}_{\mathbf{x}_{w} \in \mathcal{W}}\text{ } p\left ( y = c_{i} \mid \mathbf{x}_{w}, \mathcal{F}_{O} \right ),
\end{equation}
where $p\left ( \cdot \right )$ is the class posterior probability. This is repeated for all target classes and the class transition matrix for the web-data is estimated as: 
% \begin{equation}
% T = \left [ t_{ij} \right ]_{i,j=1}^K \text{ where }t_{ij} = p\left ( y = c_{i} \mid \hat{\mathbf{x}}_{w,c_{i}}, \mathcal{F}_{O} \right ).
% \end{equation}

\begin{equation}
T_{ij} = p\left ( y = c_{j} \mid \hat{\mathbf{x}}_{w,c_{i}}, \mathcal{F}_{O} \right )
\end{equation}
The aforementioned approach globally estimates the noise transition matrix across the web-crawled images and allows for selective diffusion across confounding classes associated with that bag. As the availability of a perfect oracle network is highly unlikely in real-world, we use a deep network trained on the limited clean dataset $\mathcal{C}$ as a potential surrogate for $\mathcal{F}_{\mathcal{O}}$. 

\noindent
\textbf{Webly Supervised Learning}: We adopt a transfer-learning like paradigm to train our fine-grained classification network as shown in Fig. \ref{fig:overview}. From an overall perspective, the web-crawled dataset $\mathcal{W}$ is used to train an initial model $\mathcal{F}_{\mathcal{W}}\left ( \cdot \right )$ with weighted-cross entropy loss. Noise correction is modulated by changing the network predictions with the estimated noise transition matrix $T$ from the retrieved web-images $\mathcal{W}$, the modulated cross-entropy loss for training $\mathcal{F}_{\mathcal{W}}$ is estimated as: 

%$T = \left [ t_{ij} \right ]_{i,j=1}^K$
% \begin{equation}
% \mathcal{L} \left (y = c_{i}, p(y \mid x \right) = - \sum_{n=1}^{N}\sum_{\mathbf{x}_{w}^{n} \in \mathcal{W}^{n}} \sum_{c=1}^{K} w_{c}\text{ }y_{\mathbf{x}_{w}^{n},c} \text{ log}\left ( t_{c,\left ( \cdot \right )} \times p\left ( c \mid \mathbf{x}_{w}^{n} \right ) \right )
% \label{eq:crossentropy}
% \end{equation}

\begin{equation}
\mathcal{L} = - \sum_{\mathbf{x}_{c}^{n} \in \mathcal{W}} w(\mathbf{x}_{c}^{n}) \ y(\mathbf{x}_{c}^{n}) \ log(T \ \times \ p(\mathbf{x}_{c}^{n}))
% \mathcal{L} = - \sum_{n=1}^{N}\sum_{\mathbf{x}_{w}^{n} \in \mathcal{W}^{n}} \sum_{c=1}^{K} w_{c}\text{ }y_{\mathbf{x}_{w}^{n},c} \text{ log}\left ( t_{c,\left ( \cdot \right )} \times p\left ( c \mid \mathbf{x}_{w}^{n} \right ) \right )
\label{eq:crossentropy}
\end{equation}
% Notice that having  $T$ as the identity matrix, would be equivalent to disable the noise correction layer. 
where $w(\mathbf{x}_{c}^{n})$ is the weight associated with the class, estimated using median-frequency balancing, $y(\mathbf{x}_{c}^{n})$ is the ground truth of sample $\mathbf{x}_{c}^{n}$ and $p(\mathbf{x}_{c}^{n})$ provides the estimated probability of sample  $\mathbf{x}_{c}^{n}$ to belong to class $c$. The trained network $\mathcal{F}_{\mathcal{W}}$ is used an initialization for subsequent fine-tuning with clean data $\mathcal{C}$ to obtain the final target model $\mathcal{F}_{\mathcal{C}}\left ( \cdot \right )$. Such a training strategy ensures that all available rich information from web-supervision is transferred as a strong prior to $\mathcal{F}_{\mathcal{C}}$ and that only expert annotated data is used to train the final network.

\section{Experiments}

% \begin{table}[t] 
% \begin{center}
% \resizebox{\textwidth}{!}{
% \begin{tabular}{c|l|l|c|c|c|c}
%  & \multirow{ 3}{*}{\textbf{Name}}  &  \multicolumn{3}{c}{Model Learning} &  \multicolumn{2}{|c}{Performance} \\ 
% \cline{3-7}
% \textbf{\#} &  &  \makecell{Training \\ Data} & Initialization & \makecell{Noise \\Correction}  & \makecell{Average \\ Accuracy} & \makecell{Cohen's \\ Kappa} \\ \hline
% \textbf{1} & BL1 & Clean & ImageNet & -  & 0.7125                & 0.6504                \\ \hline
% \textbf{2} &  & Search by Keyword & ImageNet & \checkmark  \\ 
% \textbf{3} & BL2 & Clean & \#2 & -  \\ \hline
%  \textbf{4} &  & Search by Image & ImageNet & $\times$ & & \\
% \textbf{5} & BL3 & Clean & \#4 &    & 0.7991                & 0.7602                \\ \hline
% \textbf{6} &  & Search by Image & ImageNet & \checkmark & & \\
% \textbf{7} & Proposed & Clean & \#6 & clean & 0.8053                & 0.7677                \\ \hline
% \end{tabular}
% }
% \caption{Design parameters and average performance observed for incremental baselines designed to validate WSL for skin lesion classification.}
% \label{tab:settings}
% \end{center}
% \end{table}

\textbf{Dataset:} The limited manually annotated dataset was sourced from the Dermofit Image Library~\cite{dermofit}, which consists of 1300 high quality skin lesion images annotated by expert dermatologists. The lesions are categorized into ten fine-grained classes including melanomas, seborrhoeic keratosis, basal cell carcinomas, \textit{etc.} The dataset has an extreme class imbalance (\textit{e.g.} the melanocytic nevus (25.4 \%) \textit{vs.} pyogenic granuloma (1.8\%)). The under-representation of these classes further motivates the need for augmentation with web-crawled data. For our experiments, we performed a patient-level split and used 50\% of the data for training and the rest for testing. It must be noted that due to the proprietary nature of the Dermofit library, we do not expect any of our test-data to be freely accessible via the web and hence would not be duplicated within the web-data while training the networks.

% The dataset used for our experiments is the Dermofit Image Library \cite{dermofit}, which consists of 1300 high quality skin lesion images annotated by expert dermatologists. The lesions are classified in ten different classes including melanomas, seborrhoeic keratosis, basal cell carcinomas, \textit{etc.}. A significant reason we need to increase the amount of training data we possess is that the dataset suffers from severe class imbalance. For our experiments we performed patient-level split and used 50\% of the data for training and 50\% for testing.

\noindent\textbf{Networks:} To demonstrate the contributions in terms of both the effectiveness of the proposed search by image as well as the introduction of noise correction while model learning, we established three baselines as presented in Table~\ref{tab:settings}. Specifically, BL1 is the \textit{vanilla} version of training exclusively with the clean training dataset, while contrasting with BL2 we can test the hypothesis that creating a web-dataset through \textit{search by image} induces higher search specificity and significantly reduces the cross-domain noise compared to the web data mined with keywords or user tags. We chose to use the Inception V3 deep architecture~\cite{inceptionv3} as the base model for this work. All the aforementioned models were trained with stochastic gradient descent with a decaying learning rate initialized at 0.01, momentum of 0.9 and dropout of 0.8 for regularization and the code was developed in TensorFlow \cite{tensorflow}.

% In all trained network, data augmentation is leveraged to further extend the variability in the dataset. Such augmentation include, cropping, rotation and other transformation, all randomly generated in training time.
% \begin{table}[t] 
% \begin{center}
% \resizebox{\textwidth}{!}{
% \begin{tabular}{c|l|l|c|c|c|c}
%  & \multirow{ 3}{*}{\textbf{Name}}  &  \multicolumn{3}{c}{Model Learning} &  \multicolumn{2}{|c}{Performance} \\ 
% \cline{3-7}
% \textbf{\#} &  &  \makecell{Training \\ Data} & Initialization & \makecell{Noise \\Correction}  & \makecell{Average \\ Accuracy} & \makecell{Cohen's \\ Kappa} \\ \hline
% \textbf{1} & BL1 & Clean & ImageNet & -  & 0.7125                & 0.6504                \\ \hline
% \textbf{2} &  & Search by Keyword & ImageNet & \checkmark  \\ 
% \textbf{3} & BL2 & Clean & \#2 & -  \\ \hline
%  \textbf{4} &  & Search by Image & ImageNet & $\times$ & & \\
% \textbf{5} & BL3 & Clean & \#4 &    & 0.7991                & 0.7602                \\ \hline
% \textbf{6} &  & Search by Image & ImageNet & \checkmark & & \\
% \textbf{7} & Proposed & Clean & \#6 & clean & 0.8053                & 0.7677                \\ \hline
% \end{tabular}
% }
% \caption{Design parameters and average performance observed for incremental baselines designed to validate WSL for skin lesion classification.}
% \label{tab:settings}
% \end{center}
% \end{table}
\begin{table}[t] 
\begin{center}
\resizebox{0.8\textwidth}{!}{
\begin{tabular}{c|l|l|c|c|c|c}
 & \multirow{ 3}{*}{\textbf{Name}}  &  \multicolumn{3}{c}{\textbf{Model Learning}} &  \multicolumn{2}{|c}{\textbf{Performance}} \\ 
\cline{3-7}
\textbf{\#} &  &  \makecell{Training \\ Data} & Initialization & \makecell{Noise \\Correction}  & \makecell{Average \\ Accuracy} & \makecell{Cohen's \\ Kappa} \\ \hline
\textbf{1} & BL1 & Clean & ImageNet & -  & 0.713                & 0.650                \\ \hline
 \textbf{2} &  & Web data & ImageNet & $\times$ & & \\
\textbf{3} & BL2 & Clean & \#4 &    & 0.799                & 0.760                \\ \hline
\textbf{4} &  & Web data & ImageNet & \checkmark & & \\
\textbf{5} & Proposed & Clean & \#6 &  & \textbf{0.805}                & \textbf{0.768}                \\ \hline
\end{tabular}
}
\caption{Design parameters and average performance observed for incremental baselines designed to validate WSL for skin lesion classification.}
\label{tab:settings}
\end{center}
\vspace{-32pt}
\end{table}

\section{Results and Discussion}
\begin{wrapfigure}[16]{r}{0.4\textwidth}
\vspace{-32pt}
\begin{center}
\includegraphics[width=0.18\textwidth]{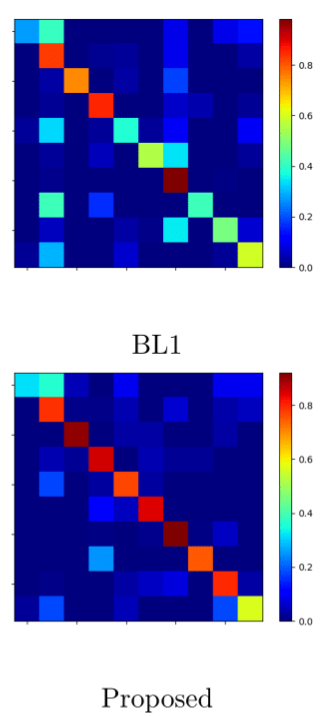}
% \caption{\small{Confusion matrices showing the effect of WSL approach compared to BL1. Significant performance is observed in WSL.}}
\caption{\small{Confusion matrices showing the effect of the proposed WSL approach compared to BL1.}}
\label{fig:confinception}
\end{center}
\end{wrapfigure}

To evaluate the effect of inclusion of WSL and the proposed noise correction, we report the average accuracy across all classes and the Cohens Kappa coefficient in Table~\ref{tab:settings}. The latter metric is particularly motivated due to the presence of significant class imbalance within our dataset. We also report the class-wise area under the curve (ROC) in Table~\ref{tab:resultsauc}. The confusion matrices are visualized in Fig.~\ref{fig:confinception}. To contrast the learned intermediate features, we embed them into a two-dimensional subspace using t-Stochastic Neighbor Embedding (t-SNE) illustrated in Fig.~\ref{fig:embedding}.  

% our method we report the area under the curve (AUC) for each class, along with the average accuracy across classes and the Cohens Kappa coefficient. The latter is particularly interesting in our case, since our dataset is characterized by significant class imbalance. 
\noindent
\textbf{Analyzing the embedding space}: Comparing the t-SNE embeddings of the test data generated by BL1 and the proposed approach in Fig.~\ref{fig:embedding}, we observe that the embeddings in WSL approach cluster examples from the same semantic class more compactly and maintain better class-separability. Within the embedding of BL1, we notice poor separability between the less-frequently occurring classes (represented with \tikzbullet{black}{black}, \tikzbullet{orange}{orange} and \tikzbullet{gray}{gray} bullets), which is significantly improved in the embedding of WSL. We also observe that the misclassification of Pyogenic Granuloma(benign) \tikzbullet{yellow}{yellow} class into Basal Cell Carcinoma (malignant) \tikzbullet{green}{green} in case of BL1 is not observed for WSL. This is quite critical as these classes are mutually exclusive. Meaning that, a vast malignant samples can be classified as benign, leading to a wrong diagnosis.

\noindent
\textbf{Effect of Web Supervision}: 
Contrasting BL1 against the proposed method in Table~\ref{tab:settings} and Fig.~\ref{fig:confinception}, we clearly observe a significant improvement in the model performance across all classes, with a more pronounced diagonal in its confusion matrix. This is clearly attributed to a better network initialization derived through transfer learning with web-supervision. This also demonstrates that we are effective in factoring out the cross-domain and cross-category noise within the web-dataset and effectively use it for supervising deep models in the presence of limited manual annotations. In Table~\ref{tab:resultsauc}, contrasting the class-wise performance, we observe that the performance on under-represented classes is significantly improved upon WSL. This is clearly evident in Intraepithelial Carcinoma \tikzbullet{gray}{gray} (5.99 \% Clean data) and Pyogenic Granuloma \tikzbullet{yellow}{yellow} (1.17 \% Clean data) where the performance improves by 3.6 \% and 7.3\% respectively. 
Contrasting BL2 with the proposed approach in Table~\ref{tab:resultsauc}, we observe an overall improvement when performing noise correction. The AUC has a slight improvement across the majority of classes. With this observation, it can be concluded that the web-crawled images retrieved in a search by image proposed methodology are so rich in terms of visual features that the effect of noise correction is only marginal when comparing BL2 and the proposed approach.

\begin{table}[t]
\centering
\resizebox{\textwidth}{!}{%
\begin{tabular}{c|c|c|c|c|c|c|c|c|c|c|c}
                  & C1              & C2              & C3              & C4              & C5              & C6              & C7              & C8              & C9              & C10             & Avg AUC         \\ \hline
Class \% in train & 3.42            & 18.49           & 4.96            & 7.53            & 5.99            & 5.82            & 25.51           & 1.17            & 19.86           & 6.67            & -               \\ \hline
BL1               & 0.898          & 0.943          & 0.976          & \textbf{0.983}          & 0.936          & 0.955          & 0.979          & 0.927          & 0.933          & 0.872          & 0.940          \\ \hline
BL2               & 0.873         & 0.955          & 0.995         & 0.966          & 0.967          & 0.984          & \textbf{0.987} & 0.991 & \textbf{0.975} & 0.935         & 0.963         \\ \hline
Proposed               & \textbf{0.920} & \textbf{0.966} & \textbf{0.995} & 0.968 & \textbf{0.972} & \textbf{0.985} & 0.983          & \textbf{0.991}          & 0.961          & \textbf{0.957} & \textbf{0.970} \\ \hline
\end{tabular}}
\caption{\small{Results showing the AUC for each class and the average overall accuracy for every model.}}
\label{tab:resultsauc}
\vspace{-25pt}
\end{table}

\section{Conclusions}
In this work, we have demonstrated for the first time the effectiveness of webly supervised learning for the task of skin lesion classification. We demonstrate that WSL can be very effective for training in limited data regimens with high-class imbalance as web data can augment under-represented classes and boost the model performance. 
By crawling the web through search by image to generate the web-dataset, we induce high search specificity and effectively minimize the influence of cross-domain noise. The proposed noise correction approach by modeling cross-category noise helps in learning an effective network initialization from web data. 

\noindent
\textbf{Acknowledgements.} The authors gratefully acknowledge CONACYT for the financial support.
This research has been funded by the Bayerische Forschungsstiftung (BSF).

% in skin lesion classification. The proposed method to retrieve data from the web has shown to reduce cross-domain noise, which is a challenging problem in webly supervised learning. In the same manner,  leveraging a noise correction approach to learn under label noise settings has resulted in the best classification performance in the ten-class skin categorization task.

\end{document}